\DocumentMetadata{
      pdfversion=2.0,pdfstandard=ua-2,
      testphase={phase-III,firstaid,math,title}
    }
    \documentclass[sigconf,screen]{acmart-tagged}
 \usepackage{makecell} % Include the makecell package
 \usepackage{soul}
\usepackage{balance}
\AtBeginDocument{%
  }
\setcopyright{cc}
\setcctype{by}
\acmDOI{10.1145/3776734.3794506}
\acmYear{2026}
\copyrightyear{2026}
\acmISBN{979-8-4007-2321-6/2026/03}
\acmConference[HRI Companion '26]{Companion Proceedings of the 21st ACM/IEEE International Conference on Human-Robot Interaction}{March 16--19, 2026}{Edinburgh, Scotland, UK}
\acmBooktitle{Companion Proceedings of the 21st ACM/IEEE International Conference on Human-Robot Interaction (HRI Companion '26), March 16--19, 2026, Edinburgh, Scotland, UK}
\received{2025-12-08}
\received[accepted]{2026-01-12}
\usepackage{xcolor,colortbl} % importing this package for table cell background colour
\usepackage{acronym}
\usepackage[inline]{enumitem}
\usepackage{listings}
\newacro{spHRI}{social-physical human-robot interaction}
\newacro{HCI}{human-computer interaction}
\newacro{HRI}{human-robot interaction}
\newacro{LLM}{large language model}
\newacro{SLM}{small language model}
\begin{document}

%\title{Charting the Growth of Social-Physical HRI (spHRI): \\ \large{A Systematic Review Pipeline Augmented by Small Language Models}}
\title[Charting the Growth of spHRI]{Charting the Growth of Social-Physical HRI (spHRI): A Systematic Review Pipeline Augmented by Small Language Models}

\author{Mayumi Mohan}
\orcid{0000-0002-0732-4476}
\authornote{denotes that both authors contributed equally to this research.}
\affiliation{%
  \institution{MPI for Intelligent Systems}
  \city{Stuttgart}
  \country{Germany}
}
\email{maymohan@is.mpg.de}

\author{Ju-Hung Chen}
\orcid{0009-0001-4756-1441}
\authornotemark[1]
\affiliation{%
  \institution{Case Western Reserve University}
  \city{Cleveland}
  \country{USA}
}
\email{jxc2350@case.edu}

\author{Alexis E. Block}
\orcid{0000-0001-9841-0769}
\affiliation{%
  \institution{Case Western Reserve University}
  \city{Cleveland}
  \country{USA}
}
\email{alexis.block@case.edu}

% \author{Mayumi Mohan}
% \authornote{denotes that both authors contributed equally to this research.}
% \affiliation{%
%   \institution{Haptic Intelligence Department\\ MPI for Intelligent Systems}
%   \city{Stuttgart}
%   \country{Germany}}
% \email{maymohan@is.mpg.de}

% \author{Ju-Hung Chen}
% \affiliation{%
%   \institution{SaPHaRI Lab \\ Human Fusions Institute\\ Case Western Reserve University}
%   \city{Cleveland}
%   \state{OH}
%     \email{ju-hung.chen@case.edu}

% \author{Alexis E. Block}
% \affiliation{%
%   \institution{SaPHaRI Lab\\ Human Fusions Institute \\ Case Western Reserve University}
%   \city{Cleveland}
%   \state{OH}
% \email{alexis.block@case.edu}

% \author{Mayumi Mohan$^{1*}$, Ju-Hung Chen$^{2*}$, and Alexis E. Block$^2$}
% \email{maymohan@is.mpg.de, {ju-hung.chen, alexis.block}@case.edu} 
% \affiliation{
%     \institution{\textit{$^1$Haptic Intelligence Department}, Max Planck Institute for Intelligent Systems, Stuttgart, Germany}
%   \city{}
%   \state{}
%   \country{}
% \institution{\textit{$^2$SaPHaRI Lab, Human Fusions Institute}, Case Western Reserve University, Cleveland, OH, USA}

\renewcommand{\shortauthors}{Mohan et al.}





\begin{abstract}
Social-physical human-robot interaction (spHRI) has grown rapidly across robotics, human-computer interaction, human-robot interaction, and haptics. Yet, fragmented terminology and inconsistent methodologies make systematic synthesis difficult. To support scalable review practices, we evaluated the extent to which small language models (SLMs; $\leq 1.5B$ parameters) can assist with title and abstract screening for a large spHRI systematic review. While no SLMs matched human reviewers' performance, the models operated locally and screened papers orders of magnitude faster. The combined SLM ensemble identified 39 papers reviewers missed, representing $10.29\%$ of the final relevant dataset. These results demonstrate that SLMs can augment, rather than replace, expert reviewers and make large-scale literature reviews accessible and sustainable.\looseness-1

% (Need to rewrite)
% Due to the technological advance and aging of the population, Human-robot interaction (HRI) is one of the most important research topics nowadays. Human has the ability to build robot that can perform either verbal or non-verbal interactions with human. However, there is a new immerge research topic of HRI that is under explored, social physical human-robot interaction (spHRI). In order to find relevant works, two researchers used a search query and downloaded most relevant publications on major databases. However, to extract the most relevant work can be a time consuming task. Thus, this study used small language models (SLMs) to reduce the work load and lower the time spend on the title and abstract screening process. In this late breaking report, this study evaluated the potential of adapting different types of SLMs on the screening process. 
\end{abstract}

\received{8 December 2025}

\maketitle

\section{Introduction}
%Research in robotics is expanding rapidly~\cite{RT1-ravichandran2025research}. 
%The field of social robotics, in particular, has experienced notable growth in recent years~\cite{RT2-mejia2017bibliometric, RT3-wang2023research}. 

\begin{figure}[b]
\vspace{-0.5cm}
    \centering
    \includegraphics[width=0.98\columnwidth, alt={A line and bar chart showing the number of social-physical human-robot interaction (spHRI) papers published annually from 1992 to 2025. Gray bars represent the count of papers per year, increasing from near zero in the early 1990s to more than 80 papers by 2025. A red line with circular markers traces the yearly counts, showing fluctuations but an overall upward trend. A smooth purple curve overlays the data, indicating a Generalized Additive Model fit with a rising trajectory and shaded confidence band. Text on the plot reports R-squared of 0.969 and Generalized Cross-Validation (GCV) of 31.756.}, trim={0cm 0cm 0 0.8cm},clip]{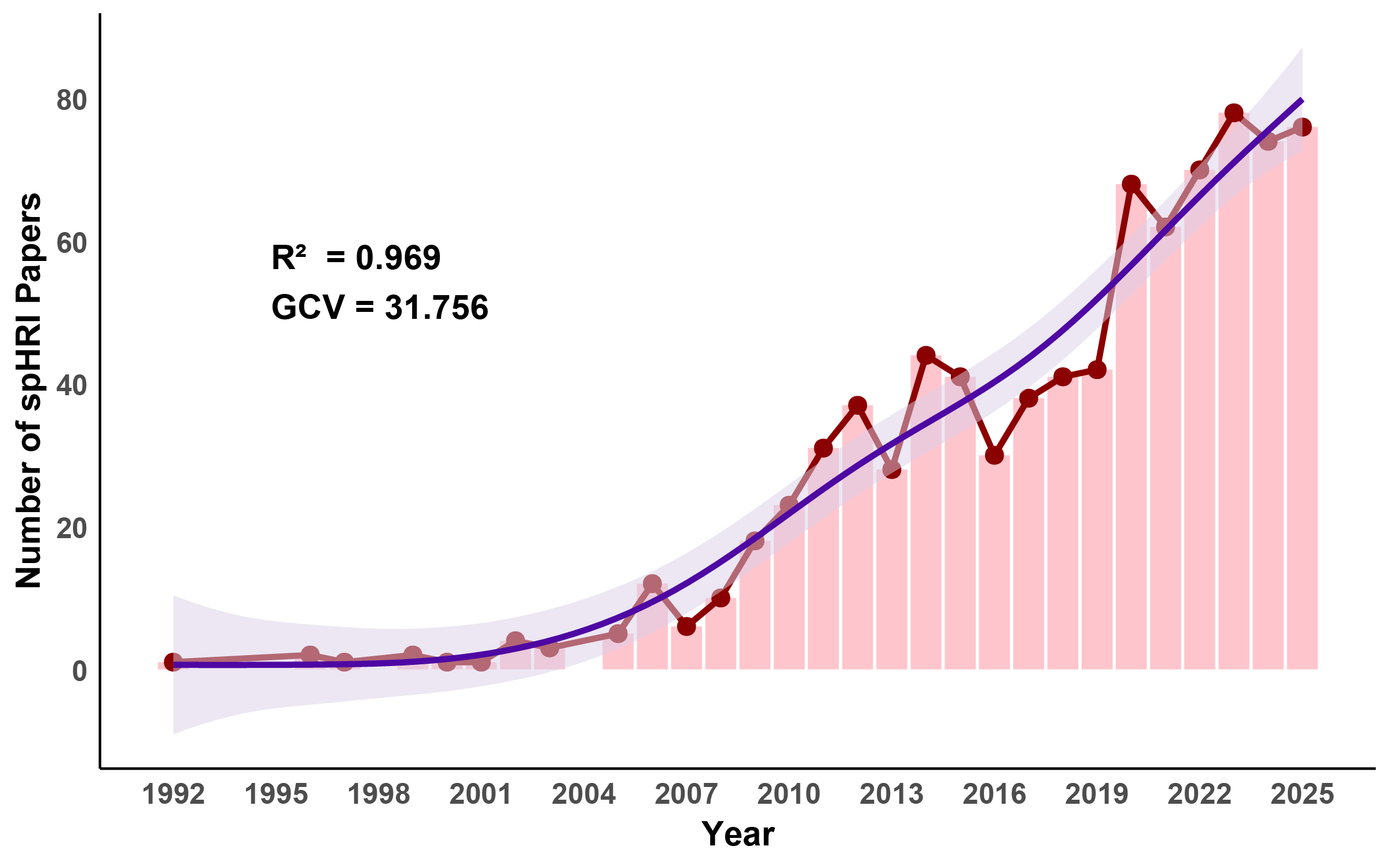}
    \Description{A line and bar chart showing the number of social-physical human-robot interaction (spHRI) papers published annually from 1992 to 2025. Gray bars represent the count of papers per year, increasing from near zero in the early 1990s to more than 80 papers by 2025. A red line with circular markers traces the yearly counts, showing fluctuations but an overall upward trend. A smooth purple curve overlays the data, indicating a Generalized Additive Model fit with a rising trajectory and shaded confidence band. Text on the plot reports R-squared of 0.969 and Generalized Cross-Validation (GCV) of 31.756}
    \vspace{-0.55cm}
    \caption{Yearly publication counts for social-physical human–robot interaction (spHRI) research from 1992–2024. The pink bars show the number of included papers per year, and the red dotted line represents the raw annual counts. A generalized additive model (GAM; purple solid line) was fitted to estimate the underlying trend, with the shaded region indicating the 95\% confidence interval. }
    \label{fig:years}
\end{figure}
Social-physical human-robot interaction (spHRI), a subfield that studies \textit{physical interactions with robots where social intent is the primary goal}, has gained growing attention~\cite{RT2-mejia2017bibliometric, RT3-wang2023research} in robotics, \ac{HRI}, \ac{HCI}, and haptics communities~\cite{spHRI8-van2015social,spHRI10-tsirka2025touch,spHRI11-paterson2023social,spHRI13-shiomi2020survey,spHRI9-huisman2017social} (see Fig.\ref{fig:years}). These interactions include both physical modalities such as affective touch, haptic feedback, and embodied emotional expression and non-tactile modalities, including gaze and verbal communication. However, despite its relevance, the \ac{spHRI} domain suffers from \textit{fragmented terminology} and \textit{disciplinary silos}, limiting systematic knowledge accumulation and comparative analysis across studies. The literature reflects this fragmentation, as \ac{spHRI}-related topics are approached from diverse disciplinary perspectives~\cite{NVC2-saunderson2019robots}, resulting in a lack of unified vocabulary across domains such as \ac{HRI}, \ac{HCI}, affective computing, and healthcare. The absence of consistent terminology complicates efforts to compare studies, classify interaction types, or interpret results consistently~\cite{NVC2-saunderson2019robots, spHRI8-van2015social, spHRI9-huisman2017social, spHRI13-shiomi2020survey}. For example, a survey on nonverbal communication in \ac{HRI} extensively examines the impact of robot touch, yet does not explicitly reference the \ac{spHRI} construct~\cite{NVC2-saunderson2019robots}. 

Existing surveys often focus predominantly on direct interpersonal touch with robots and overlook interactions with robotic objects (robjects), physical interfaces, wearables, or haptic devices~\cite{NVC2-saunderson2019robots, spHRI13-shiomi2020survey, spHRI10-tsirka2025touch, spHRI9-huisman2017social}, resulting in gaps in how \ac{spHRI} is understood and classified. To address these gaps, we initiated a large-scale systematic literature review,
%with the goal of articulating clearer conceptual foundations and proposing guidelines for future \ac{spHRI} research. %Third, the broader \ac{HCI} literature on touch, which includes rich work on tactile communication, attachment objects, and haptic design, is rarely brought into conversation with \ac{HRI}-centered reviews, even though many \ac{HCI}systems function as precursors or analogs to social robots.
drawing from over 130,000 candidate publications across robotics, \ac{HRI}, \ac{HCI}, and related venues. The review aimed to clarify conceptual foundations and map the methodological landscape of \ac{spHRI} research. However, the scale of modern literature presents a critical bottleneck for human reviewers. Comprehensive title and abstract screening, an early but essential stage, can be prohibitively time-consuming. Large Language Models (LLMs) are increasingly being explored for this task, but most require cloud access, 
%However, state-of-the-art LLMs are not sustainable and often not accessible. First, they are highly resource-intensive, requiring 
substantial computational power~\cite{SUSTAIN1-strubell2019energy}, raise ethical concerns due to carbon footprint and propriety opacity~\cite{SUSTAIN2-bender2021dangers}, or accessibility issues due to prohibitive costs
%. Third, due to their size and complexity, many modern LLMs require expensive, high-performance hardware, such as multi-GPU or TPU clusters, making them inaccessible to many researchers and institutions
~\cite{SUSTAIN2-bender2021dangers}. 

Thus, in this Late-Breaking Report (LBR), we investigate how open-source, lightweight small language models (SLMs), which run locally on consumer hardware, can support accessible and environmentally responsible systematic reviews as a supportive screening aid without sacrificing accuracy. Specifically, we ask:
\textbf{To what extent can an SLM serve as a reliable second-pass filter that recovers missed relevant papers in title and abstract screening?} We do not frame SLMs as replacements for human reviewers. The performance comparisons are used to assess the reliability of SLMs. To answer our research question, we benchmark SLMs against human reviewers and test their utility, precision, conservatism, and potential role as assistive screeners in large-scale review workflows by testing the following hypotheses:
\vspace{-\topsep}
\begin{itemize}[nosep]
    % \item [\textbf{H1:}] Each evaluated \ac{SLM} will demonstrate screening performance (e.g., precision, recall, and F1 score) comparable to prior published evaluations in similar NLP tasks.
    \item [\textbf{H1:}] spHRI exhibits sustained publication growth consistent with an emergent subfield.
    %A PRISMA-guided title and abstract screening will reveal sustained growth of social-physical HRI (spHRI) publications over time, demonstrating that spHRI exhibits the bibliometric markers of an emerging HRI subfield.
    %A PRISMA-guided title and abstract screening of the HRI literature will identify at least $T$ eligible studies that meet our \ac{spHRI} inclusion criteria, indicating that there is sufficient volume and coherence of work for \ac{spHRI} to be treated as its own subfield.
    \item [\textbf{H2:}] At least one \ac{SLM} will perform comparably to human reviewers in classifying abstracts as relevant or irrelevant, with no statistically significant difference in classification outcomes at $\alpha = 0.05$.
    \item [\textbf{H3:}] A well-performing \ac{SLM} will identify at least 5\% of relevant papers missed by human reviewers during the title and abstract screening, suggesting its use as a secondary reviewer.
    % \item [\textbf{H3:}] Compared to human reviewers, \ac{SLM}s will produce a significantly higher false negative rate during title and abstract screening, indicating a more conservative inclusion strategy. 
\end{itemize}
\vspace{-0.7cm}
\section{Related Work}
% \vspace{-0.2cm}
Prior work in the \ac{HRI}~\cite{NVC2-saunderson2019robots}, \ac{HCI}~\cite{spHRI8-van2015social} and haptics~\cite{spHRI9-huisman2017social} communities has explored various forms of physical interaction between humans and robots. However, these works often treat physical contact as a modality, rather than a \textit{distinct research domain}. As a result, \ac{spHRI} remains \textit{conceptually underdeveloped}. Only recently have researchers begun to focus more explicitly on \ac{spHRI}~\cite{spHRI13-shiomi2020survey, spHRI11-paterson2023social}. Foundational surveys in haptics \cite{spHRI9-huisman2017social} and nonverbal robot communication~\cite{NVC2-saunderson2019robots} discuss robot-initiated or received touch, but rarely define it in \textit{explicit social or relational terms}.
%remains primarily grounded in haptic technology and rehabilitation contexts, and therefore does not frame social-physical interaction with robots as a central theme. 

More recent reviews have attempted to address this limitation. For example,~\citet{spHRI13-shiomi2020survey}, surveyed various touch modalities in \ac{HRI} but provided only \textit{high-level taxonomies}. \citet{spHRI10-tsirka2025touch} offered a comprehensive systematic review of robot touch, including emotional outcomes, but stopped short of proposing a \textit{unifying theoretical framework} for \ac{spHRI}. Key elements such as directionality (who initiates touch), morphology (robot vs. object), and purpose (affective vs. functional) remain inconsistently defined across studies. Moreover, interactions with robotic interfaces, wearable systems, and robjects are often excluded, limiting generalizability and taxonomy development. At the same time, the scale of modern scholarly output poses a practical barrier to comprehensive literature synthesis. Traditional systematic reviews, though methodologically rigorous, require \textit{substantial human effort} and are often outdated by the time of publication. Thus, researchers have begun using large language models (LLMs) to assist with different stages of the systematic review pipeline~\cite{LLMSR27-bayani2025leveraging, LLMSR7-lieberum2025large}. The most common applications include title/abstract screening~\cite{LLMSR1-castillo2023leveraging,LLMSR2-thode2025exploring,LLMSR5-khraisha2024can}, full-text review~\cite{LLMSR5-khraisha2024can}, data extraction~\cite{LLMSR5-khraisha2024can}, question formulation, search strategy refinement~\cite{LLMSR10-marassi2025comparing}, summary generation~\cite{LLMSR12-yun2023appraising}, and scientific writing. Additionally, many commercially available AI-based tools exist to support literature reviews~\cite{LLMSR3-silva2025enhancing}. GPT-based models, in particular, are widely used due to their generalizability and high accuracy \cite{LLMSR4-sujau2025accelerating,LLMSR5-khraisha2024can, LLMSR13-felizardo2024chatgpt, LLMSR6-zimmermann2024leveraging}.

However, many LLMs are cloud-based, computationally expensive, or environmentally unstable \cite{SUSTAIN2-bender2021dangers, SUSTAIN1-strubell2019energy}, posing accessibility and ethical concerns. As an alternative, SLMs have emerged as a lightweight, locally deployable solution.  While many open-source LLMs exist, most are still moderately sized (7B+ parameters), often requiring specialized hardware to deploy efficiently~\cite{qin2025empirical}. This computational overhead remains a barrier for many \ac{HRI} research labs. Thus, we investigate SLMs with $1.5$ billion parameters or fewer, which can run locally on consumer-grade desktops and enable more accessible, scalable, and sustainable review processes.
%prioritize efficiency and low computation cost at the expense of performance. 
%The most commonly used LLMs are those from the GPT family~\cite{LLMSR4-sujau2025accelerating,LLMSR5-khraisha2024can}, which consistently outperform other systems in screening and data extraction tasks~\cite{LLMSR13-felizardo2024chatgpt}. GPT models are also frequently used as baselines for comparison~\cite{LLMSR6-zimmermann2024leveraging}. 
%Several studies have evaluated open-source alternatives such as LLaMA and Mixtral~\cite{LLMSR2-thode2025exploring,LLMSR4-sujau2025accelerating}. 
Though fewer studies have benchmarked SLMs in systematic review contexts~\cite{LLMSR2-thode2025exploring, LLMSR4-sujau2025accelerating}, early evidence suggests models like LLaMA and Mixtral can perform competitively \cite{LLMSR2-thode2025exploring,LLMSR4-sujau2025accelerating, LLMSR5-khraisha2024can, LLMSR13-felizardo2024chatgpt}. Still, most existing benchmarks focus on biomedical or software engineering domains, and not \ac{HRI}.
%Llama-2 variants generally performed the best, though they still were behind GPT-based models GPT-4~\cite{LLMSR5-khraisha2024can, LLMSR13-felizardo2024chatgpt}. 

%To our knowledge, \textit{no prior work} has systematically evaluated the use of \ac{SLM}s for screening publications in \ac{spHRI} or broader \ac{HRI} contexts. Nor have such tools been tested for their ability to detect subtle, socially meaningful patterns in physical interaction studies. By addressing the \textit{conceptual fragmentation} in \ac{spHRI} and the \textit{methodological bottleneck} of literature screening, this work advances both domains. 
To the best of our knowledge, no prior work has conducted a \textit{comprehensive systematic review} of \ac{spHRI} that spans \ac{HRI}, \ac{HCI}, haptics, and robotics. Given the volume, diversity, and conceptual nuance of this literature, we argue that \ac{spHRI} merits recognition as a \textit{distinct subfield} within HRI, with its own frameworks, taxonomies, and design challenges. At the same time, the growing scale of interdisciplinary research calls for more \textit{accessible, efficient, and transparent} review methods. While LLMs have begun to support systematic reviews in other domains, SLMs remain underevaluated in HRI contexts. Their potential to reduce reviewer burden and democratize systematic reviews makes them an important tool for expanding inclusive research practices in our field.
\textit{Taken together, prior surveys do not deliver a unified conceptual or methodological account of spHRI, nor do they scale to contemporary publication volume. This work addresses both gaps.}

% ... explain difference llm vs slm.

% llm as secondary

% existing

%(need to check references)
%LLM has been used in the review process and review paper writing process. In the most case, LLM was used for screening, data extraction and writing.

%However, there are some issues of using LLM during the review process, including inaccurate citation, unnatural wording and incorrect fabricated information. Thus, using LLM as a second reviewer during the title and abstract screening has gain popularity. 

%(includes publications about title and abstract screening by LLM)

%Despite the popularity and efficiency of using LLM, most of the work are running the LLM on cloud or high power computing center. For some researchers with lack of fund or access to high computing power machines, title and abstract screening become an time consuming process. Thus, this study present a method by using LLMs with small parameter size and run the LLM locally.

% \subsection{Prompting LLMs for Systematic Reviews}
% - Who has compared with other llms only?
% - comparison with llms and human?

 \vspace{-0.1cm}
\section{Methods}
We followed a preregistered \cite{BlockMohanChen2025_spHRI_OSF} PRISMA workflow~\cite{page2021prisma} and evaluated four locally deployed SFqwenLMs (individually and as a unanimity ensemble) as a second-pass aid for title/abstract screening (Fig.~\ref{fig:prompt})%This section outlines the procedures for our systematic literature review and the implementation of the \ac{SLM}-based screening pipeline. We assume a workflow in which human reviewers first screen titles and abstracts using standard inclusion and exclusion criteria. Next, multiple SLMs act as secondary screeners and their predictions are aggregated into a conservative ensemble. Papers flagged by the ensemble are subsequently manually re-screened by researchers.\looseness-1 
 % \vspace{-0.3cm}
 \subsubsection*{\textbf{Review Methodology}}
% \vspace{-0.2cm}
%Our review followed PRISMA guidelines\cite{page2021prisma} and was preregistered on the Open Science Framework [Reference Anonymized].
%Our review was preregistered on the Open Science Framework and followed PRISMA guidelines~\cite{page2021prisma}. 
%Our goal was to identify, map, and analyze research in social physical human-robot interaction (spHRI) literature to (1)~establish clear definitions for the field, (2)~identify the scope and characteristics of existing research, (3)~summarize methodological approaches, and (4)~articulate future directions and research gaps. 

\begin{figure}
    \centering
    \includegraphics[width=1\linewidth, alt={This figure presents a left-to-right workflow combining human screening and small language model (SLM)–assisted re-screening. On the left side, a PRISMA-style flow shows record reduction and human screening. A large initial pool of studies is reduced first by automated duplicate removal and then by researcher screening. Two independent researchers each screen half of a 32,000-record subset. After manual removal of a small number of additional duplicates, most records are judged irrelevant and a small subset is judged relevant. These human relevance decisions form the input to the next stage. A vertical dotted line visually separates human screening (left) from SLM-assisted analysis (right). On the right side, both the human-included and human-excluded sets are evaluated by four small language models (Gemma, DeepSeek, Qwen, and Llama), shown as running on a desktop computer. Their outputs are combined into an SLM ensemble. The ensemble produces two outcomes: (1) a group of studies where the ensemble agrees with the human inclusion decisions, and (2) a larger group of studies that humans had excluded but the ensemble flags as potentially relevant. Only the flagged ``potentially missed'' studies are sent back to a researcher for re-screening. A small fraction of these are reclassified as relevant and added to the set moving forward. The workflow converges on a final box labeled ``full text review,'' which combines the originally human-included studies with the additional studies recovered through SLM-assisted re-screening. The diagram emphasizes that SLMs are used as a secondary safety net to detect potentially missed relevant studies, rather than replacing human screening.}]{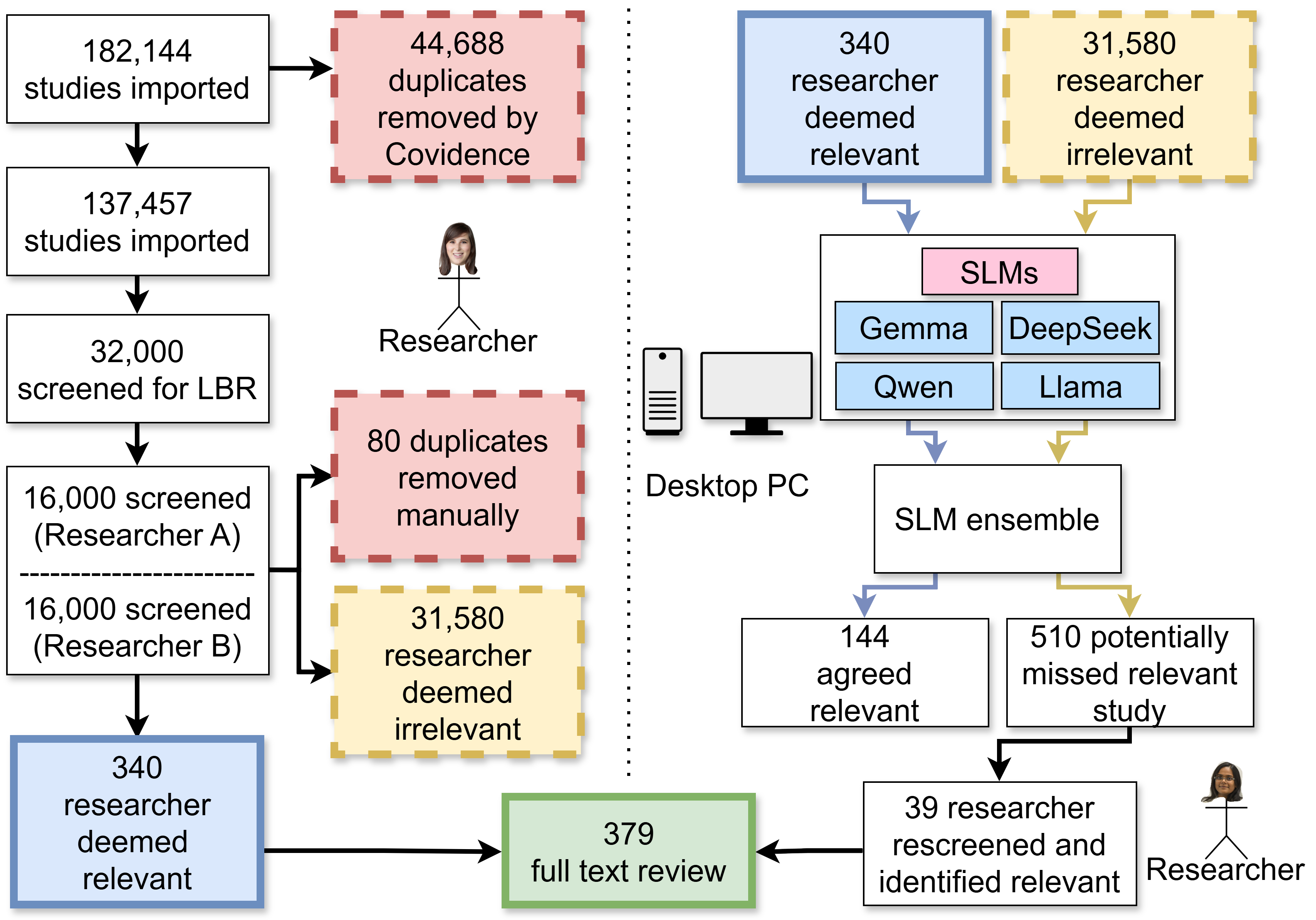}
    \Description{This figure presents a left-to-right workflow combining human screening and small language model (SLM)–assisted re-screening. On the left side, a PRISMA-style flow shows record reduction and human screening. A large initial pool of studies is reduced first by automated duplicate removal and then by researcher screening. Two independent researchers each screen half of a 32,000-record subset. After manual removal of a small number of additional duplicates, most records are judged irrelevant and a small subset is judged relevant. These human relevance decisions form the input to the next stage. A vertical dotted line visually separates human screening (left) from SLM-assisted analysis (right). On the right side, both the human-included and human-excluded sets are evaluated by four small language models (Gemma, DeepSeek, Qwen, and Llama), shown as running on a desktop computer. Their outputs are combined into an SLM ensemble. The ensemble produces two outcomes: (1) a group of studies where the ensemble agrees with the human inclusion decisions, and (2) a larger group of studies that humans had excluded but the ensemble flags as potentially relevant. Only the flagged ``potentially missed'' studies are sent back to a researcher for re-screening. A small fraction of these are reclassified as relevant and added to the set moving forward. The workflow converges on a final box labeled ``full text review,'' which combines the originally human-included studies with the additional studies recovered through SLM-assisted re-screening. The diagram emphasizes that SLMs are used as a secondary safety net to detect potentially missed relevant studies, rather than replacing human screening.}
    \vspace{-0.75cm}
    \caption{Overview of the PRISMA and SLM-assisted screening workflow. For this late-breaking report (LBR), 32,000 of 182,144 imported papers were screened by two independent researchers (16,000 each, no overlap). The researchers manually removed 80 additional duplicates, identified 340 papers as relevant, and excluded 31,580 as irrelevant. Both the included and excluded sets were then evaluated on a desktop PC using four small language models (SLMs; Gemma, DeepSeek, Qwen, and Llama) individually and as an ensemble. The SLM ensemble agreed with the researchers' inclusion decisions for 144 papers and flagged 510 excluded records as potentially missed relevant studies. These records were re-screened by a researcher, yielding 39 additional relevant papers. In total, 379 studies advanced to full-text review.}
    \vspace{-0.4cm}
    \label{fig:prompt}
\end{figure}

We defined eligible systems as physically embodied robots or agents that exhibit autonomous, reprogrammable, or responsive behavior for socially meaningful physical interactions. These systems included humanoid robots, therapeutic robotic animals, mobile service robots, and wearable robotic devices. We excluded passive or non-embodied systems lacking autonomous control, feedback, or human-agent physical interaction. Records were retrieved from  PubMed, IEEE Xplore, Scopus, ACM Digital Library, and major academic databases spanning robotics/HRI/HCI/haptics venues. 
%We also included English-language gray literature. 
The final search query was refined and tested with seed papers (e.g.~\cite{block_softness_2019,lc_sit_2024,stiehl2005design}) to capture spHRI-relevant studies. The query included the following boolean terms: \textit{(robot OR agent OR bot OR object OR wearable) AND (touch OR haptic OR tactile OR physical OR embodied) AND (social OR affective OR emotional OR trust OR motivation OR engagement OR empathy OR support)}. %To ensure query quality, we \textit{iteratively refined the string} and verified that key seed papers were consistently returned across their respective databases. Since several databases returned extremely large, relevance-ranked result sets, retrieval was feasibility-limited. 
Given extremely large relevance-ranked result sets, we exported records in descending relevance until yield declined, defined as 500 consecutive out-of-scope record from rapid title/abstract screening (e.g., IEEE Xplore: 21,270 $\rightarrow$ 5,499 exported). Prior reviews in related fields began with roughly 1,000 candidate papers~\cite{spHRI9-huisman2017social,spHRI10-tsirka2025touch}. 
%We verified that key seed papers were consistently retrieved across databases during query refinement.% (936 sphri9 and 139 (sphrispHRI10-tsirka2025touch)
% papers than most systematic reviews. 
Search dates ranged from June 2, 2025 to October 20, 2025. 
%vOur search covered the following databases: PubMed, IEEE Xplore, ACM Digital Library, Scopus, Web of Science, EBSCOhost, Taylor \& Francis Online, SpringerLink, Wiley Online Library, Nature, Science, arXiv, and PsyArXiv. In addition, we collected papers from major robotics, HRI, and HCI conferences such as ICRA, IROS, HRI, CHI, ICSR, RO-MAN, as well as available workshop proceedings.
%earch engines often produced large numbers of irrelevant results; therefore, for each database, we collected all papers until reaching a 
All retrieved records were imported into Covidence \cite{babineau2014product}, which removed 44,687 duplicates automatically. Dataset size, de-duplication, and labeling procedures are summarized in Fig.~\ref{fig:prompt}; we treat yes/maybe as relevant and no as irrelevant for evaluation.

% For this LBR, we evaluate the 32,000 records (of 137,457 total papers) for title and abstract screening (two researchers, 16,000 each; no overlap). Each paper was rated as ``yes,'' ``no,'' or ``maybe'' fitting the inclusion/exclusion criteria. 
%All ``maybe'' and ``yes'' papers advanced to full-text review. 
%During this process, they identified and removed an additional 80 duplicates. Our SLM evaluation uses the 32,000 records screened at the title/abstract stage. We treat ``yes/maybe'' as \textit{relevant} and ``no'' as \textit{irrelevant} when computing performance metrics.\looseness-1
% \vspace{-0.1cm}
 \subsubsection*{\textbf{\ac{SLM} Reviewer Implementation}}
% \vspace{-0.2cm}
To assess the extent to which SLMs can support early-stage screening, we evaluated four open-source, lightweight \ac{SLM}: Llama3.2 (1B parameters), Gemma3 (1B), Qwen3 (0.6B) and DeepSeek-R1 (1.5B). These models were selected for their small parameter sizes and ability to run locally without cloud infrastructure. All models were run locally using the Ollama framework~\cite{marcondes2025using} on a commercial desktop computer equipped with an Intel i5-13600K CPU, NVIDIA RTX4070 GPU (12GB VRAM) and 32 GB of system RAM. We developed a Python script to interact with the Ollama API. The script sent the model a structured prompt describing the review objective, inclusion/exclusion criteria, and paper title and abstract.  We also provided the model with the seed papers that had been used to refine the search query. Each model then returned a binary include/exclude decision. All SLMs were executed with their default parameter settings. 

%The prompt was designed to communicate the objective, query string, and inclusion and exclusion criteria to the \ac{SLM}. We also provided the model with the seed papers that had been used to refine the search query. Since we are conducting a preliminary exploratory study we did not provide the \ac{SLM} with a system prompt but focused on developing a good user prompt. Thus, we prompted the \ac{SLM} to find papers fulfilled the following criteria: \textit{1)~Objective:} Identify papers on social-physical interaction between humans and embodied agents. \textit{2)~Query String:} (robot OR agent OR bot OR object OR wearable) AND (touch OR haptic OR tactile OR physical OR embodied) AND (social OR affective OR emotional OR trust OR motivation OR engagement OR empathy OR support) \textit{3)~Inclusion Criteria:} Must involve social touch as a key part of the study; agent must be physically embodied; morphology and touch directionality do not matter. \textit{4)~Exclusion Criteria:} No physical interaction; purely verbal or emotional interaction; no social intent; agent not embodied; no human involvement.
 \subsubsection*{\textbf{\ac{SLM} Reviewer Implementation}}
We aggregate predictions from the four SLMs using a unanimity (set-intersection) rule: a paper is flagged only if all SLMs predict ``Yes''. %Equivalently, the flagged set is the intersection of the four models' inclusion sets. 
This ensemble rule is intentionally conservative, mitigating false-positives from individual models with high false positive rates (FPR) and keeping the manual re-screening subset small. The two human reviewers then re-screened the flagged subset, each independently screening $50\%$ of the flagged records. No additional post-processing is applied beyond this unanimity rule.
\subsubsection*{\textbf{Statistical Analysis}}
% \vspace{-0.2cm}
%All SLM-based screening was implemented in Python. 
To evaluate the performance of SLMs, we calculated the average completion time per paper (in seconds), the proportion of true positives (TP), true negatives (TN), false positives (FP), and false negatives (FN). \textit{The average completion time} (seconds per paper) was computed by dividing the total screening runtime by the number of papers processed. For the Combined \ac{SLM} condition, we summed the average per-paper completion times of each individual \ac{SLM} and the overhead of the MATLAB script per paper. %This script imported a CSV of all screened papers and evaluated agreement across models. Its average runtime was calculated by dividing the total execution time by the number of papers analyzed. A \textit{true positive} occurred when an \ac{SLM} correctly identified a paper as relevant; a \textit{true negative} when it correctly identified a paper as irrelevant; a \textit{false positive} when it incorrectly labeled an irrelevant paper as relevant; and a \textit{false negative} when it failed to identify a relevant paper. Each value was normalized by the total papers screened.
From these quantities, we calculated several standard classification metrics. These included accuracy, recall, false positive rate (FPR), precision, specificity and F1 score. These metrics were first calculated relative to the initial human labels and then recomputed using the updated reference set obtained from re-screening the 510 ensemble-flagged records.%\textit{Accuracy} measures the proportion of all correct classifications, computed as $(TP + TN)/(TP + TN + FP + FN)$. \textit{Recall}, also known as \textit{sensitivity} or \textit{true positive rate}, represents the proportion of actual positives correctly identified and is defined as $TP/(TP + FN)$. The \textit{false positive rate (FPR)} quantifies the proportion of actual negatives that were incorrectly classified as positive, calculated as $FP/(FP + TN)$. \textit{Precision} measures the proportion of predicted positives that are correct and is defined as $TP/(TP + FP)$.
%\textit{Specificity} measures the true negative rate, i.e., the proportion of actual negatives correctly identified, given by $TN/(TN + FP)$.
%Finally, the \textit{F1 score} provides the harmonic mean of precision and recall and is given by $(2 \cdot Precision \cdot Recall)/ (Precision + Recall)$. 
%We first computed these metrics for each SLM comparing each SLM's binary predictions to the initial human reviewer screening labels. We then recomputed all metrics relative to the updated reference set after manual re-screening of the 510 ensemble SLM flagged records.% In Table.~\ref{table:rescreen}, ``Human reviewer'' refers to the \textit{original single-pass human screening decisions} evaluated against the updated reference set, i.e., counting the 39 recovered papers as false negatives for the initial human screen.

Statistical inference used paired nonparametric bootstrapping ($10,000$ resamples). Metrics were recomputed for the human reviewer and each model on every resample. Model–human differences were evaluated using two-sided bootstrap tests with $95\%$ confidence intervals. Holm–Bonferroni correction was applied for multiple comparisons, with $\alpha = 0.05$.

% \begin{itemize} [nosep]
%     \item \textbf{Accuracy:} The proportion of all correct positive or negative classifications ($(TP+TN)/(TP+TN+FP+FN)$).
%     \item \textbf{Recall:} The true positive rate or \textbf{sensitivity} the proportion of correctly classified actual positives ($(TP+TN)/(TP+TN+FP+FN)$).
%     \item \textbf{False Positive Rate:} Proportion of positively classified actual negatives  classifications ($TP/(FP+TN)$).
%     \item \textbf{Specificity:}The true negative rate or proportion of correctly classified negatives ($TN/(TN+FP)$).
% \end{itemize}

% \subsection{Prompt}

% Below are positive examples (SPHRI = Yes):

% \{Candidate Papers title and abstract\}

% Now classify the new publication:

% Title: \{title\}
% Abstract: \{abstract\}

% Does this belong to Social Physical Human-Robot Interaction (SPHRI)?
% Answer with only "Yes" or "No".

% - Keywords
% - Be liberal
% - Candidate papers

% \subsection{Random Maybe papers}
% - robotic barstools
% - human drone interaction
% - human touching robot
% - how to handle conference proceeding

%For the purposes of this review, we define a robot as any \textit{artificially engineered, physically embodied system capable of sensing the environment and interacting with the world in a reprogrammable and intentional manner.} We are particularly interested in robots or agents that are physically embodies and can interact with world in the service of \textit{social goals}. We thus include humanoid robots and non-anthropomorphic platforms such as therapeutic robotic animals (e.g., Paro \cite{hung2019benefits}), mobile service robots (e.g., Chairbot \cite{lc2024sit, agnihotri2019persuasive}), and wearable robotic devices (e.g., haptic vests, sleeves, exoskeletons) when they exhibit \textit{autonomy, responsiveness, or reprogrammable behavior} directed toward human-centered outcomes. Importantly, we exclude purely passive devices or garments lacking autonomous control or feedback. Instead, our focus is on systems that exhibit at least minimal agency or responsiveness in a physical and socially meaningful context, aligning with the broader goals of spHRI."

\section{Results}
% \vspace{-0.2cm}

\subsubsection*{H1: Evidence of an emergent subfield}
To assess whether \ac{spHRI} exhibits the publication dynamics characteristic of an emerging research subfield, we analyzed the yearly distribution of included papers identified through our PRISMA-guided screening. Figure~\ref{fig:years} shows a clear upward trajectory in publication volume from $1992$ to $2025$. A generalized additive model (GAM) was fitted to the data to capture the underlying trend. The fitted smooth term revealed sustained nonlinear growth, with a pronounced acceleration over the past decade. The GAM accounted for $96.9\%$ of the variance in yearly publication counts ($R^2 = 0.969$) and yielded a generalized cross-validation score of $31.756$, indicating strong model fit and a non-random pattern of increase. 
\vspace{-0.2cm}
\subsubsection*{H2: Performance compared to human reviewer} 
\begin{figure}
    \centering
    \includegraphics[width=0.98\columnwidth, alt={A grid of six pie charts showing the proportion of papers predicted as relevant versus irrelevant by four small language models, Gemma, DeepSeek, Qwen, Llama, and a Combined SLM, and by a human reviewer. Each pie chart has a dark solid segment for predicted positives and a light dotted segment for predicted negatives. Gemma shows the largest dark segment, predicting almost all papers as relevant. DeepSeek and Qwen show smaller dark segments. Llama shows a much smaller predicted-positive segment. The Combined SLM shows a very narrow dark segment. The human reviewer's pie chart shows the smallest predicted-positive area, with nearly the entire chart representing predicted negatives.}, trim={0cm 0cm 0 0},clip]{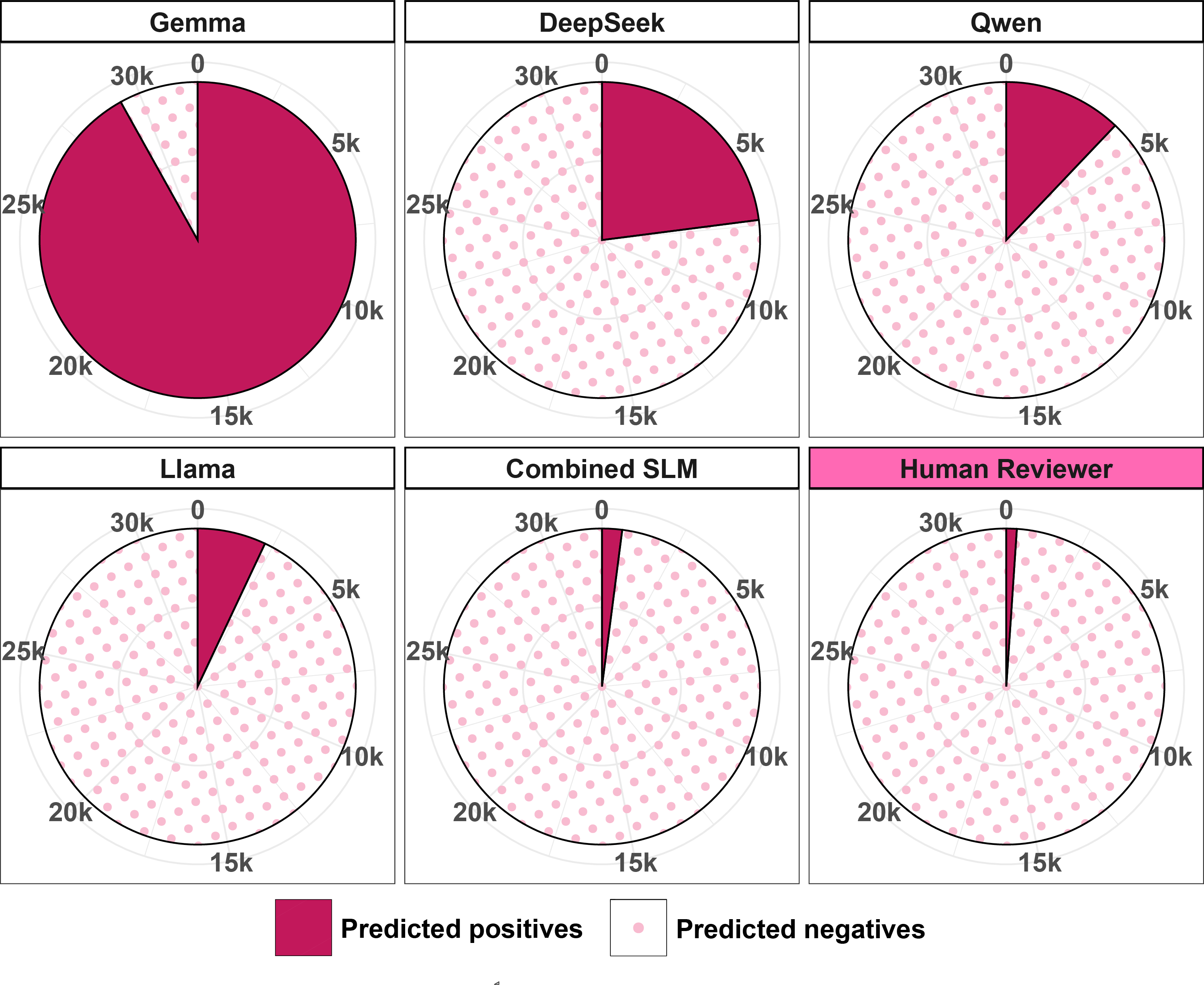}
    \Description{A grid of six pie charts showing the proportion of papers predicted as relevant versus irrelevant by four small language models, Gemma, DeepSeek, Qwen, Llama, and a Combined SLM, and by a human reviewer. Each pie chart has a dark solid segment for predicted positives and a light dotted segment for predicted negatives. Gemma shows the largest dark segment, predicting almost all papers as relevant. DeepSeek and Qwen show smaller dark segments. Llama shows a much smaller predicted-positive segment. The Combined SLM shows a very narrow dark segment. The human reviewer’s pie chart shows the smallest predicted-positive area, with nearly the entire chart representing predicted negatives.}
    \vspace{-0.5cm}
    \caption{Classification outcomes of each \ac{SLM}, the combined SLM ensemble, and the human reviewer, showing the proportion of papers predicted positively as relevant (magenta, solid fill) versus those that were predicted negatively (light pink, dotted). Each pie chart represents a sample of 31,920 screened titles and abstracts.} 
    \label{fig:pie}
    \vspace{-0.5cm}
\end{figure}

Table~\ref{table:initial} reports performance metrics for four individual SLMs, a Combined \ac{SLM} ensemble, and a human reviewer. All five models performed significantly worse than the human reviewer (Fig.~\ref{fig:pie}) across all evaluation metrics (paired nonparametric bootstrap, Holm-adjusted $p < 0.001$ for all comparisons). However, Llama achieved the best standalone agreement with the human labels ($specificity=0.9366$, $FPR = 0.0634$) while screening abstracts in $0.21$ seconds per item, orders of magnitude faster than human review (30–60 seconds/paper). The Combined \ac{SLM} reduced the false positive rate to $0.00161$, outperforming all individual models. However, there is a pronounced recall–workload tradeoff. For example, Gemma achieves very high recall ($0.9794$) but with extreme over-inclusion ($FPR =0.9185$,$ precision=0.0114$), whereas the Combined \ac{SLM} is far more selective ($precision=0.2202$) at the cost of recall ($0.4235$).

%The SLM-based screening provided a mixed result of the screening process. For screening 31920 publications, Llama only takes 0.21 seconds to screen a single publication on average, which is significantly faster than other SLMs. Followed by Gemma (0.39 seconds/paper). DeepSeek had the slowest completion time (1.75 second/paper), while Qwen had a completion time of 0.93 second/paper.

\begin{table}[hb!]
\vspace{-0.5cm}
\caption{Performance metrics for automated title and abstract screening across four SLMs and the combined \ac{SLM}, evaluated against the initial human screening labels (single-screen, no overlap). Reported values include average completion time, classification outcomes (TP, TN, FP, FN), and derived performance metrics (accuracy, recall, false positive rate, precision, specificity, and F1 score). Since initial human labels serve as the reference here, derived metrics not defined for the human reviewer and are reported as N/A.} 
\begin{center}
\resizebox{0.47\textwidth}{!}{
\begin{tabular}{c c c c c c c c} 
 \hline
 Item & Gemma & DeepSeek & Qwen & Llama & \makecell{Combined \\ SLM} &  \makecell{Human \\ Reviewer}\\ [0.5ex] 
 \hline\hline
Avg. completion time \\ (seconds/paper)& 0.39 & 1.75 & 0.93 & 0.21 & 3.2801 & 30-60\\
 \hline
 True Positive & 0.0104 & 0.0070 & 0.0074 & 0.0073 & 0.0045 & N/A\\%0.0107 \\
 \hline 
 True Negative & 0.0807 & 0.7667 & 0.8758 & 0.9266 & 0.9734 & N/A\\%0.9893 \\
 \hline
 False Positive & 0.9087 & 0.2227 & 0.1136 & 0.0628 & 0.0160 & N/A \\
 \hline
 False Negative & 0.0002 & 0.0037 & 0.0033 & 0.0034& 0.0061  & N/A \\
 \hline
 \hline
 Accuracy & 0.0911 & 0.7737 & 0.8832 & 0.9339 & 0.9779 & N/A \\
 \hline
 Recall & \cellcolor{gray!20} \textbf{0.9794} & 0.6559 & 0.6941 & 0.6853 & \cellcolor{gray!20} \textbf{0.4235} & N/A \\
 \hline
 FPR & 0.9185 & 0.2251 & 0.1148 & \cellcolor{gray!20} \textbf{0.0634} & 0.00161 & N/A \\
 \hline
 Precision & \cellcolor{gray!20} \textbf{0.0114} & 0.0304 & 0.0611 & 0.0104 & \cellcolor{gray!20} \textbf{0.2202} & N/A \\
 \hline
 Specificity & 0.0815 & 0.7749 & 0.8852 & \cellcolor{gray!20} \textbf{0.9366} & 0.9839 & N/A \\
 \hline
 F1 Score & 0.0224 & 0.0581 & 0.1124 & 0.1809 & 0.2897 & N/A \\
 \hline \hline
\end{tabular}
}
\vspace{-0.3cm}
%\caption{Performance metrics for automated title and abstract screening across four SLMs and the combined \ac{SLM}, evaluated against the initial human screening labels (single-screen, no overlap). Reported values include average completion time, classification outcomes (TP, TN, FP, FN), and derived performance metrics (accuracy, recall, false positive rate, precision, specificity, and F1 score). Since initial human labels serve as the reference here, derived metrics not defined for the human reviewer and are reported as N/A.} %Llama achieved the best balance of speed and overall accuracy, screening papers in 0.21 seconds per item. Gemma exhibited high recall but low specificity due to frequent false positives.%, while DeepSeek and Qwen relative to Llama.}}
\label{table:initial}
\vspace{-0.2cm}
\end{center}
\end{table}

\begin{table}[h!]
% \vspace{-0.5cm}
\begin{center}

\vspace{-0.5cm}
\caption{Revised performance metrics after manual re-screening of 510 papers flagged by the Combined \ac{SLM}. This additional screening identified 39 relevant papers missed during the initial human review, resulting in a corrected ground truth for true positive and false positive calculations. All metrics were recalculated using this updated baseline. ``Human reviewer'' refers to the \emph{original single-pass human screening decisions} evaluated against the updated baseline.} 
\resizebox{0.47\textwidth}{!}{
\begin{tabular}{c c c c c c c c} 
 \hline
 Item & Gemma & DeepSeek & Qwen & Llama & \makecell{Combined \\ SLM} &  \makecell{Human \\ Reviewer}\\ [0.5ex] 
 \hline\hline
 True Positive & 0.0116 & 0.0082 & 0.0086 & 0.0085 & 0.0057 & 0.0094 \\
 \hline 
 True Negative & 0.0807 & 0.7667 & 0.8758 & 0.9266 & 0.9734 & 0.9893 \\
 \hline
 False Positive & 0.9075 & 0.2215 & 0.1123 & 0.0615 & 0.0148 & 0.0012 \\
 \hline
 False Negative & 0.0002 & 0.0037 & 0.0033 & 0.0034& 0.0061  & 0 \\
 \hline
 \hline
 Accuracy & 0.0923 & \cellcolor{gray!20} \textbf{0.7749} & 0.8844 & 0.9351& 0.9791 & 0.9988 \\
 \hline
 Recall & \cellcolor{gray!20} \textbf{0.9815} & 0.6913 & \cellcolor{gray!20} \textbf{0.7256} & 0.7177 & \cellcolor{gray!20} \textbf{0.4828} & 1 \\
 \hline
 FPR & 0.9184 & \cellcolor{gray!20} \textbf{0.2241} & 0.1137 & 0.0623 &  \cellcolor{gray!20} \textbf{0.0149} & 0.0012 \\
 \hline
 Precision & 0.0127 & \cellcolor{gray!20} \textbf{0.0357} & 0.0712 & 0.1216 & \cellcolor{gray!20} \textbf{0.2798} & 0.8853 \\
 \hline
 Specificity & 0.0816 & 0.7759 & 0.8863 & 0.9377 & 0.9851 & 0.9988 \\
 \hline
 F1 Score & 0.0250 & 0.0679 & 0.1297 & 0.2080 & \cellcolor{gray!20} \textbf{0.3543} & \cellcolor{gray!20} \textbf{0.9392} \\
 \hline \hline
\end{tabular}
}
% \vspace{-0.3cm}
%\caption{Revised performance metrics after manual re-screening of 510 papers flagged by the Combined \ac{SLM}. This additional screening identified 39 relevant papers missed during the initial human review, resulting in a corrected ground truth for true positive and false positive calculations. All metrics were recalculated using this updated baseline. ``Human reviewer'' refers to the \emph{original single-pass human screening decisions} evaluated against the updated baseline.} 
%, yielding adjusted values for accuracy, recall, false positive rate, precision, specificity, and F1 score.}}
%The Combined \ac{SLM} demonstrates the highest F1 score under the revised ground truth, indicating its utility as a secondary reviewer capable of recovering papers overlooked by human screening.}}
\vspace{-0.7cm}
\label{table:rescreen}
\end{center}
\end{table}

%In terns of screening accuracy, Llama shows the closest result as researcher ground truth on screening publications, with 95.56\% of accuracy. Gemma has the highest Recall (90.00\%), indicate the ability to identify the relevant documents. However, Gemma has the lowest accuracy among all the SLMs and shows highest False Positive rate (66.63\%). Although DeepSeek and Qwen takes significantly longer time to screen publications, the result shows that DeepSeek and Qwen do not perform better than Llama in terns of accuracy. Overall, Llama shows the closest alignment as human researcher's screening result.
\vspace{-0.2cm}
\subsubsection*{H3: Performance as secondary reviewers} 
The manual rescreening of the $510$ papers flagged by the Combined \ac{SLM} revealed an additional $39$ papers that were missed during the original human review. When added to the $340$ papers initially included, these $39$ papers represent $10.29\%$ of total relevant studies ($379$), more than double our predefined $5\%$ threshold. Since relevant papers are rare in our large corpus, accuracy is dominated by true negatives and is less informative than recall or precision (Table~\ref{table:rescreen}). For example, DeepSeek appears reasonably accurate ($0.7749$) but has low precision ($0.0357$) and low FPR ($0.2241$), implying that many of its ``include'' decisions are incorrect. Based on the FPR (0.0149) and precision (0.2798), the combined \ac{SLM} is highly selective with a correspondingly lower recall (0.4828). Additionally, all SLMs screen substantially faster than manual review, enabling inexpensive secondary passes at scale.

%selective second-pass filter (low FPR, producing a small re-screening workload), and achieves the highest highest F1 score among all automated conditions. Thus, the Combines \ac{SLM} is provides a favorable precision to recall tradeoff for targeted manual follow-up screening. 

\vspace{-0.1cm}\section{Discussion and Conclusion}
This study explored the extent to which an \ac{SLM} can serve as a reliable second-pass filter that recovers missed relevant papers in title and abstract screening. Our findings show that (1)~\ac{spHRI} is a coherent and growing research area (H1 supported), (2)~\ac{SLM}s cannot yet mirror expert judgment (H2 rejected), but (3)~they meaningfully improve review pipelines (H3 supported). The results expose a precision–recall tradeoff typical of imbalanced screening: models can achieve high recall by over-including (raising false positives), while the unanimity rule suppresses false positives but sacrifices recall. Thus, SLMs are most useful as a high-throughput safety net that enables targeted manual re-screening and not replacements for human reviewers. %As \ac{spHRI} expands across platforms, modalities, and populations, scalable screening tools will become essential. 
Our workflow provides a practical, accessible pathway for integrating lightweight SLMs into the review pipeline while maintaining rigor. 

% This part is the limitation and the future work that the reviewer requested
\balance
While this LBR shows SLMs have promise in screening relevant studies in systematic reviews, it has several limitations, including binary (no-confidence) SLM outputs and single-pass human screening without overlap, which prevents inter-rater reliability analysis. Additional missed papers may exist since we only re-screened the papers flagged by the Combined \ac{SLM}. %There may be papers that were
%In this study, the SLM was only asked to answer yes or no regarding the relevance of the study; the confidence level remains unknown. Furthermore, human researcher screening without overlap affects the quality of the screened studies. The inter-rater reliability cannot be calculated, and the quality of the screened studies cannot be ensured.
Ongoing work will refine prompts (including confidence scoring), evaluate alternative SLM ensembles, and examine additional assessment metrics while expanding screening to the full dataset.

% Since this LBR only covered $23\%$ of the imported studies, we will expand and complete the abstract and title screening process by improving the prompt structure.  The SLM prompt will be improved by adding the SLM's confidence level of study relevance. Furthermore, the combination of different types of SLMs will also be evaluated on the effectiveness and accuracy of the screening process.

\textit{spHRI has outgrown scattered investigation}; its sustained growth now demands recognition and consolidation as a coherent research area. As the field expands, the volume and conceptual diversity of work make it challenging for the average researcher to correctly identify all relevant papers without substantial domain knowledge. This challenge is amplified for newcomers to a field, who lack the historical and conceptual grounding needed to navigate an emerging landscape. For example, \citet{shibata2012therapeutic}'s work on therapeutic robot touch contains no explicit reference to social touch in its title or abstract, yet an expert would immediately flag it for inclusion in full-text review based on its foundational role in physical interaction research. Here, SLMs offer tangible value: they can help researchers make sense of a fractured literature, capture missed studies, and support the systematic synthesis required to unify \ac{spHRI} into a mature, well-defined subfield.

\begin{acks}
\vspace{-0.1cm}
We thank Austin Wilson and Chitvan Killawala. This work was partially supported by the German Federal Ministry of Research, Technology and Space (BMFTR) under the Robotics Institute Germany (RIG).\looseness-1 
\end{acks}

\bibliographystyle{ACM-Reference-Format}
\bibliography{refs}


\end{document}